\documentclass[letterpaper, 10 pt, conference]{ieeeconf}  

\IEEEoverridecommandlockouts                              

\RequirePackage[2018/12/01]{latexrelease}
\usepackage{mathptmx} 
\usepackage{times} 
\usepackage{array}
\usepackage{cite}
\usepackage{balance}
\usepackage{mathtools}
\usepackage{url}
\usepackage{xcolor}
\usepackage{scalerel,stackengine}
\stackMath
\usepackage{multirow}

\newcommand\reallywidehat[1]{%
\savestack{\tmpbox}{\stretchto{%
  \scaleto{%
    \scalerel*[\widthof{\ensuremath{#1}}]{\kern-.6pt\bigwedge\kern-.6pt}%
    {\rule[-\textheight/2]{1ex}{\textheight}}
  }{\textheight}%
}{0.5ex}}%
\stackon[1pt]{#1}{\tmpbox}%
}

\title{\LARGE \bf Task-Driven Deep Image Enhancement Network for Autonomous Driving in Bad Weather}

\author{
{Younkwan Lee$^{1}$, Jihyo Jeon$^{2}$, Yeongmin Ko$^{1}$, Byunggwan Jeon$^{1}$, and Moongu Jeon$^{1,2}$} \\
\thanks{This work was partly supported by Ministry of Culture, Sports and Tourism and Korea Creative Content Agency(Project Number: R2020070004) and Institute of Information \& communications Technology Planning \& Evaluation (IITP) grant funded by the Korea Government (MSIT) (No. 2014-3-00077, Development of Global Multi-target Tracking and Event Prediction Techniques Based on Real-time Large-Scale Video Analysis), and the National Research Foundation of Korea (NRF) grant funded by the Korea Government (MSIT) (No. 2019R1A2C2087489).}%
\thanks{$^{1}$Machine Learning \& Vision Laboratory, Gwangju Institute of Science and Technology (GIST), Gwangju 61005, South Korea
        {\tt\small\{brightyoun, koyeongmin, jbk34071, mgjeon\}@gist.ac.kr}}%
\thanks{$^{2}$Korea Culture Technology Institute (KCTI), Gwangju 61005, South Korea
        {\tt\small jihyo@gm.gist.ac.kr}}
}

\begin{document}

\maketitle
\thispagestyle{empty}
\pagestyle{empty}

\begin{abstract}

    Visual perception in autonomous driving is a crucial part of a vehicle to navigate safely and sustainably in different traffic conditions.
    However, in bad weather such as heavy rain and haze, the performance of visual perception is greatly affected by several degrading effects.
    Recently, deep learning-based perception methods have addressed multiple degrading effects to reflect real-world bad weather cases but have shown limited success due to 1) high computational costs for deployment on mobile devices and 2) poor relevance between image enhancement and visual perception in terms of the model ability.
    To solve these issues, we propose a task-driven image enhancement network connected to the high-level vision task, which takes in an image corrupted by bad weather as input.
    Specifically, we introduce a novel low memory network to reduce most of the layer connections of dense blocks for less memory and computational cost while maintaining high performance.
    We also introduce a new task-driven training strategy to robustly guide the high-level task model suitable for both high-quality restoration of images and highly accurate perception.
    Experiment results demonstrate that the proposed method improves the performance among lane and 2D object detection, and depth estimation largely under adverse weather in terms of both low memory and accuracy.

\end{abstract}

\section{INTRODUCTION}

    Autonomous vehicles require comprehensive and accurate visual perception to safely navigate diverse driving conditions with little or no human effort.
    Currently, visual perception tasks are achieved by deep neural networks (DNN) which have demonstrated impressive performance on a wide range of high-level vision tasks, such as lane detection \cite{neven2018towards,ko2020key}, monocular depth estimation \cite{godard2017unsupervised,pillai2019superdepth,godard2019digging}, and scene recognition \cite{lin2017focal,9304708,miller2018dropout,yu2020unsupervised,zhou2020joint}.
    The success of deep convolutional neural networks relies on a large number of high-quality images and a large computational cost on large-scale resource devices.
    However, existing models do not typically consider the degradations taken from bad weather conditions for training as well as low-resource devices to be deployed on mobile devices.
    Therefore, one needs to train complex visual degradations caused by bad weather with DNN having lower resource consumption.
    
    
    
    \begin{figure}[t]
    \begin{center}
       \includegraphics[width=1.0\linewidth]{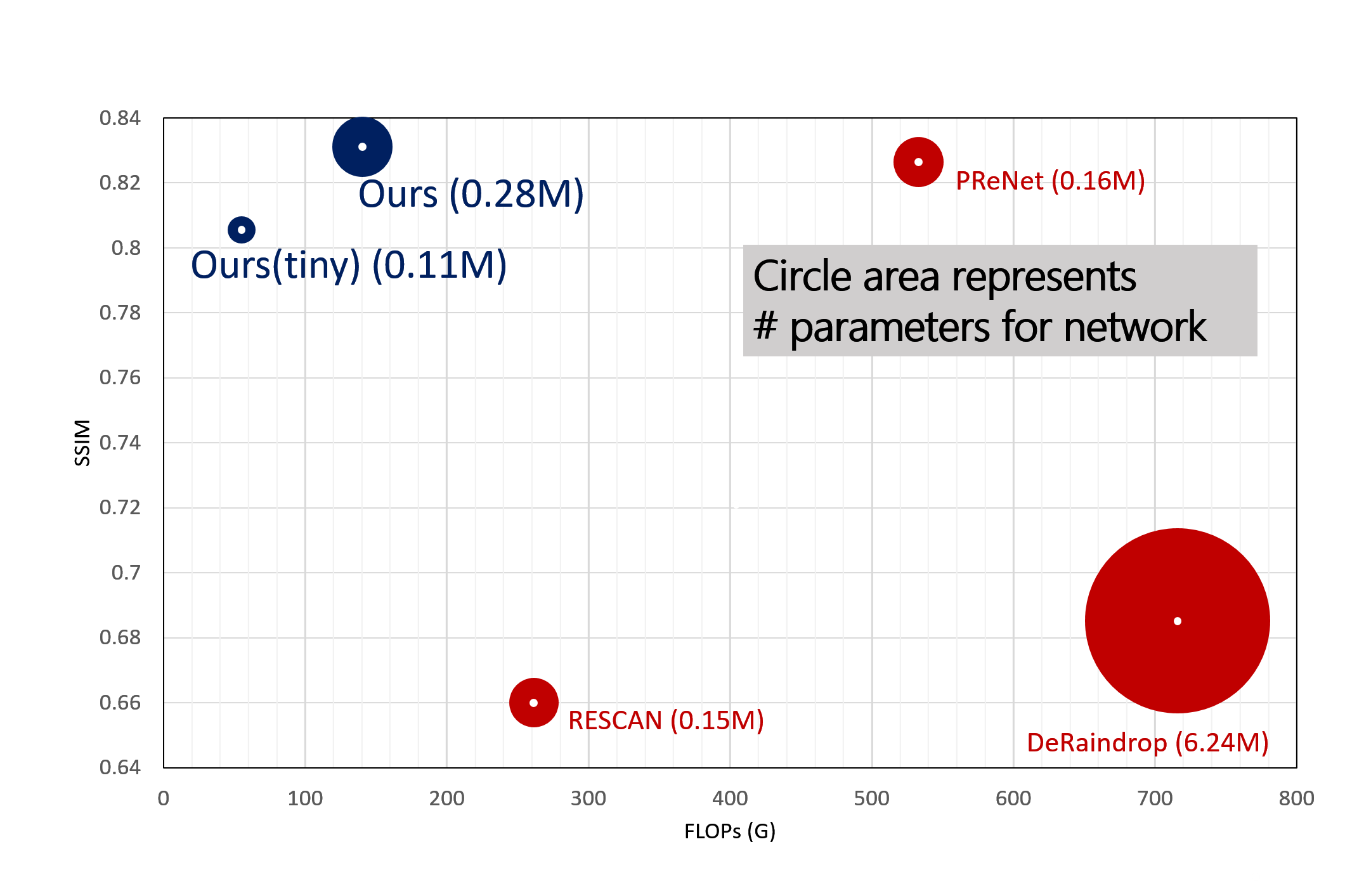}
    \end{center}
    \vspace{-0.8cm}
       \caption{\textbf{TuSimple \cite{tuSimple} SSIM vs FLOPs.} The area of the circle plot represents the parameters of the model as described in parentheses. Compared to three methods (in blue): RESCAN \cite{li2018recurrent}, DeRaindrop \cite{qian2018attentive} and PReNet \cite{ren2019progressive}, our proposed models (in red) are shown to achieve perception- and hardware-friendly performance, while reducing model size.}
    \vspace{-0.6cm}
    \end{figure}
    
    To solve this problem, image enhancement has been adopted as a key solution \cite{zhang2017beyond,ren2016single,ren2017video,li2019single,ren2019progressive,liu2020self}, enabling it to improve image quality by itself as a preprocessing method within independent components.
    Benefited from these methods, the latest deep learning-based enhancement models seem almost plausible at first glance.
    Unfortunately, however, we observe that existing enhancement methods still not suitable for high-level vision tasks in bad weather, and even worsen performance in some cases.
    Among them, we find its limitations in the following reasons.
    First, They usually rely on metrics based on the human visual system that are not correlated with visual perception models such as PSNR and SSIM.
    Thus, when the image is inferred by them, the recovered information is not sufficient or appropriate for high-level tasks.
    Second, they generally require a lot of computational power despite having to run autonomous driving with resource-limited platforms.
    Hence they are unaffordable for autonomous driving when integrated into the resource-constrained environments with other perception models.
    

    In this paper, we introduce a novel task-driven image enhancement framework that benefits by exploring the mutual influence between visual perception and enhancement for safe and reliable autonomous driving in bad weather conditions.
    To this end, our model aims to have \textit{perception- and hardware-friendly} characteristics against any bad weather situations as end-to-end learning, as shown in Fig. 1.
    To the best of our knowledge, this is the first attempt to connect image enhancement and high-level vision for autonomous driving under multiple bad weather conditions.
    In summary, our work makes following key contributions:
     \begin{itemize}
     \vspace{-0.1cm}
    \item We propose a universal multiple bad weather removal framework that enables the high-level vision tasks to improve the robustness of existing models without degradation and retraining. 
    
    \item We develop a task-driven enhancement network for less memory and computational cost, thus it is suitable for the embedded system when building ADAS (Advanced Driver Assistance System) for autonomous vehicles.
    
    \item 
    We introduce a novel training strategy that minimizes the detrimental effects of image enhancement while improving the performance of high-level tasks in an end-to-end and task-driven manner.

    \item 
    We experimentally validate the effectiveness of the proposed method when embeds high-level tasks such as lane detection, monocular depth estimation, and object detection. To our best knowledge, this work is one of few studies to apply the proposed image enhancement module for visual high-level tasks of autonomous driving under bad weather.
    \vspace{-0.3cm}
    \end{itemize}
    
\section{RELATED WORKS}
\vspace{-0.1cm}
    \subsection{Bad Weather Image Enhancement Algorithms}
    \vspace{-0.1cm}
        Many models and algorithms have been developed to deal with only one weather condition.
        For example, \cite{fu2017removing,li2018recurrent,qian2018attentive,zhang2019image,ren2019progressive} have been proposed to recover rain effects, including rain streaks or raindrops.
        In \cite{li2019stacked}, desnowing was designed with a multi-scale stacked densely connected CNN for detecting and removing snowflakes from a single snowy image.
        A few approaches for defogging/dehazing was proposed by non-local prior \cite{berman2016non} or image-to-image translation network without relying on the physical scattering model \cite{qu2019enhanced}.
        However, they are not designed and trained for all the bad weather conditions, thus may not guarantee to build safe autonomous driving in bad weather. 
        The above issue of universal bad weather enhancement has been addressed by hybrid all-in-one model \cite{li2017aod,sun2019convolutional,li2020all}.
        In \cite{sun2019convolutional}, a joint dehazing and deraining CNN was proposed with the classical atmospheric scattering model from the global context of a single image.
        In \cite{li2020all}, generative adversarial networks were used by relying on task-specific encoders that only process a particular degradation type.
        Although these all-on-one methods have achieved impressive performance on bad weather image enhancement, most of them were only suited for one specific kinds of perception task, such as object detection \cite{li2020all} or semantic segmentation \cite{sun2019convolutional} without studying various high-level tasks.
        Moreover, their methods are computationally too inefficient for on-device embedding in autonomous vehicles and are not suitable for fast inference.
        To the best of our knowledge, our work is the first study to provide faster processing time and compressed parameters while being perception- and hardware-friendly to deal with a variety of bad weather conditions.
    
    \begin{figure*}[t]
    \vspace{1.3cm}
    \begin{center}
       \includegraphics[width=0.6\linewidth]{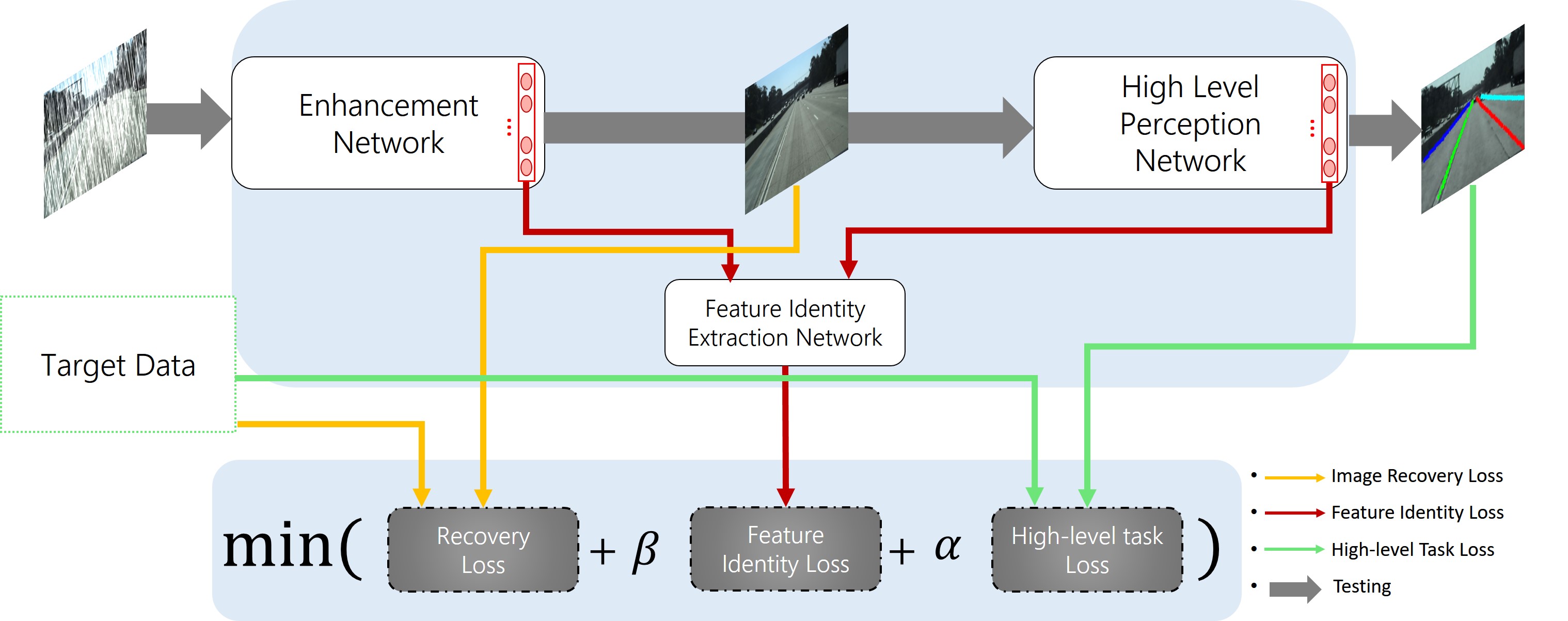}
    \end{center}
       \caption{\textbf{Overview of our proposed enhancement framework.} Our framework comprises a low memory enhancement network, a task-specific high-level perception network, and a feature identity extraction network. We connect all networks into one pipeline and train in an end-to-end manner.}
    \end{figure*}
    
    \subsection{Limitation of High-Level Vision Models}
        When high-level vision tasks are conducted in bad weather conditions that they often encounter in autonomous driving, image enhancement is usually worked as an independent pre-processing stage, which might be poorly related to the task-specific goal \cite{hasirlioglu2016test,bijelic2018benchmarking,liu2019enhance}.
        Recently, limitation of deep learning-based high-level vision models has been investigated to reveal their inefficiency against bad weather conditions that they operate with image enhancement methods as the independent pre-processing stage.
        For example, \cite{pei2018does} demonstrated that the existing image dehazing methods do not bring much benefit to help the image classification performance based on the analysis of the evaluation metric.
        Similarly, \cite{bahnsen2018rain,li2019single} showed that existing image deraining models do not much improve the performance of recognition model, or worsened, based on images collected in the real world.
        To comprehend such problem, some researches \cite{nguyen2015deep,wang2016studying,wu2017relation} pointed out that visual enhancement works mainly focus on human perception quality \cite{lai2016comparative}, which becomes harmful by visual artifact patterns or noise perturbation.
        
        Nevertheless, there are several studies to overcome the vulnerability of high-level vision models.
        In \cite{li2017aod}, re-formulated atmospheric scattering model that direct reconstructs haze-free images was studied by using an end-to-end learning scheme.
        In \cite{zendel2018wilddash}, various factors of image degradation were tackled by analyzing the semantic segmentation networks in autonomous driving scenes.
        In \cite{sun2019convolutional,Lee_2019_ICCV,son2020urie} image enhancement and high-level task were jointly designed as an end-to-end learning model, achieving improved performance over both tasks.
        However, most methods still consider some weather effects such as rain or fog, separately.
        In addition, their optimization is still not suitable for high-level visual tasks, and there is no consideration of the efficiency of the hardware.
        As far as we know, our method is the first attempt to propose a recognition- and hardware-friendly framework, taking into account various bad weather conditions.
        
    \begin{figure}[t]
    \begin{center}
       \includegraphics[width=0.8\linewidth]{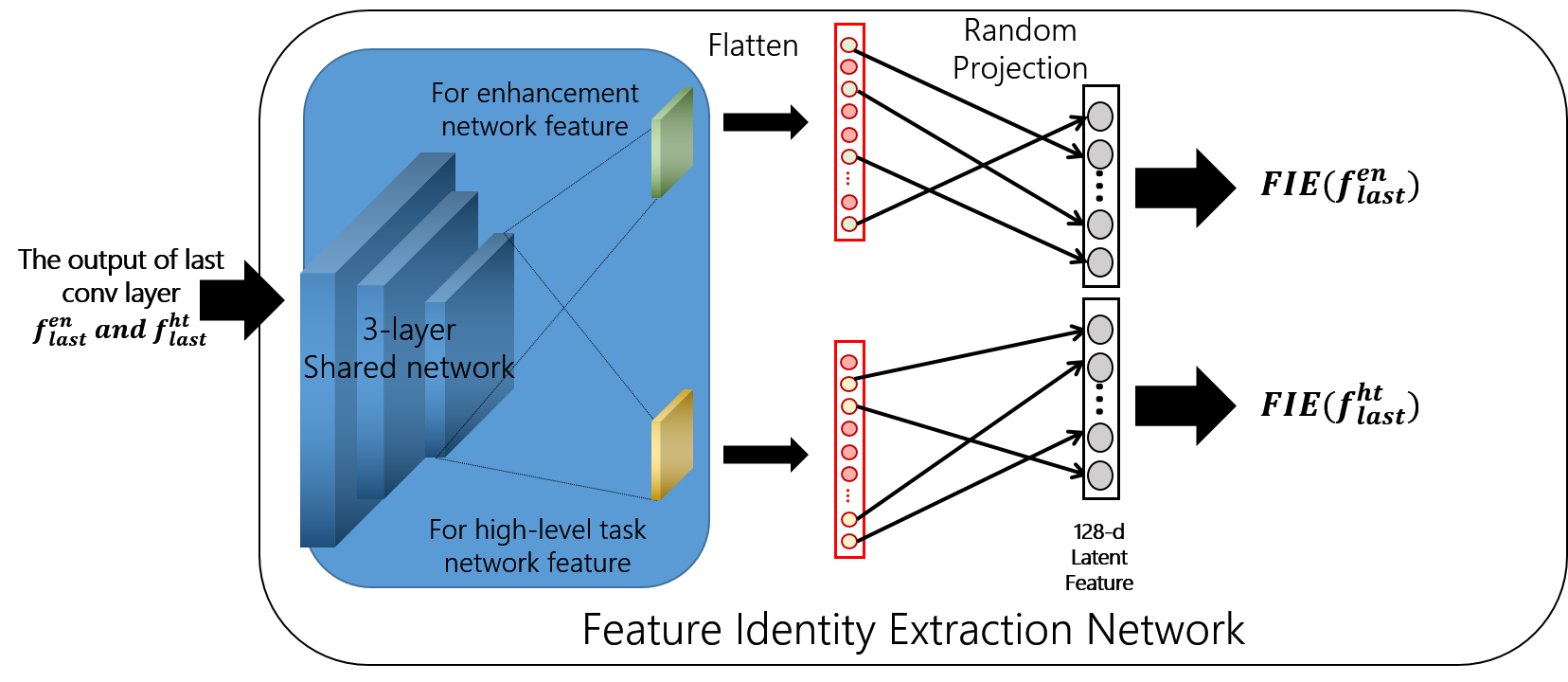}
    \end{center}
    \vspace{-0.5cm}
       \caption{\textbf{The detailed structure in the feature identity extraction network (FIE).} The random projection shows the connections to 128-d latent feature space from the last flatten representation of each network.}
    \vspace{-0.5cm}
    \end{figure}

\section{PROPOSED METHOD}

\subsection{Problem Definition} 
    Here we present the general setting of the problem prior to the illustration of the proposed method.
    We have a clean image $I^{GT}$ and corresponding bad weather image $I^X$.
    We define that both images have the same high-level task label $Y^{GT}$.
    The bad weather input $I^X$ is first fed into the image enhancement network $E^{en}$ and outputs the recovered image $I^{pred}$, while the last layer before the final output of $E^{en}$ represents $f^{en}_{last}$.
    Subsequently, the recovered image $I^{pred}$ is fed forward through the high-level perception network $E^{ht}$ and outputs the high-level perception result $Y^{pred}$ with the last convolution layer $f^{ht}_{last}$.
    The parameters of each network are represented as $\theta_{en}$ and $\theta_{ht}$ with pre-trained for each task, where $\theta_{ht}$ is frozen while optimizing the proposed method.
    Note that we do not explicitly define a detailed network for high-level task, since our proposed is applicable to arbitrary high-level task baselines.
    Additionally, the last layers of the two networks mentioned above, $f^{en}_{last}$ and $f^{ht}_{last}$, are respectively fed into feature identity extraction network with the learnable parameter $\phi$.
    Fig. 2 shows the overall framework of the networks.

\begin{table}[]
\centering
\caption{\textbf{The architecture of enhancement network (71 layers).} Info is composed of kernel and stride.}
\vspace{-0.3cm}
\label{tab:my-table}
\begin{tabular}{llll}
\hline
\textbf{ID} & \textbf{Layer/Block} & \textbf{Info} & \textbf{Output} \\ \hline
\textbf{Input} & $Input_{bad}$ & \- & \-  \\
\textbf{\begin{tabular}[c]{@{}l@{}} $Input_{init}$ \\ $Input_{x^0}$ \end{tabular}} & \begin{tabular}[c]{@{}l@{}} $Input_{x^0}$\\ $Input_{bad}$ \end{tabular} & - & - \\ \hline
\textbf{for t=1 to T} &  &  &  \\ \hline
Concat1 & $Concat(Input_{bad},Input_{x^{t-1}})$ & - & 1024x512x6 \\
Conv1 & Conv(Concat1) & 3, 1 & 1024x512x32 \\
HBlock1 & HBlock(Conv1) : 8 layers & 3, 1 & 1024x512x32 \\
Concat2 & Concat(Conv1, HBlock1) & - & 1024x512x64 \\
Conv2 & Conv(Concat2) & 1, 1 & 1024x512x32 \\
Add1 & Add(Conv1, Conv2) & -, - & 1024x512x32 \\
Conv3 & Conv(Add1) & 3, 1 & 1024x512x32 \\
HBlock2 & HBlock(Conv3) : 16 layers & 3, 1 & 1024x512x32 \\
Concat3 & Concat(Conv3, HBlock2) & - & 1024x512x64 \\
Conv4 & Conv(Concat3) & 1, 1 & 1024x512x32 \\
Add2 & Add(Conv3, Conv4) & -, - & 1024x512x32 \\
Conv5 & Conv(Add2) & 3, 1 & 1024x512x32 \\
HBlock3 & HBlock(Conv5) : 16 layers & 3, 1 & 1024x512x32 \\
Concat4 & Concat(Conv5, HBlock3) & - & 1024x512x64 \\
Conv6 & Conv(Concat4) & 1, 1 & 1024x512x32 \\
Add3 & Add(Conv5, Conv6) & -, - & 1024x512x32 \\
Conv7 & Conv(Add3) & 3, 1 & 1024x512x32 \\
HBlock4 & HBlock(Conv7) : 16 layers & 3, 1 & 1024x512x32 \\
Concat5 & Concat(Conv7, HBlock4) & - & 1024x512x64 \\
Conv8 & Conv(Concat5) & 1, 1 & 1024x512x32 \\
Add4 & Add(Conv7, Conv8) & -, - & 1024x512x32 \\
Conv9 & Conv(Add4) & 3, 1 & 1024x512x32 \\
HBlock5 & HBlock(Conv9) : 4 layers & 3, 1 & 1024x512x32 \\
Concat6 & Concat(Conv9, HBlock5) & - & 1024x512x64 \\
Conv10 & Conv(Concat6) & 1, 1 & 1024x512x32 \\
Add5 & Add(Conv9, Conv10) & -, - & 1024x512x32 \\
Conv11 & Conv(Add5) & 3, 1 & 1024x512x3 \\ \hline
\multicolumn{4}{l}{\textbf{Recursive Output: $Input_{x^t}$ (For Restoration Learning)}} \\ \hline
\textbf{end for} &  &  &  \\ \hline
\end{tabular}
\vspace{-0.6cm}
\end{table}

\subsection{Enhancement Network Architecture}
    Our network is inspired by DenseNet \cite{huang2017densely} as a feature encoding network.
    Feature encoding network has one efficient structure for high-resolution applications at the edge and outperforms existing image enhancement methods by leveraging HarDNet \cite{chao2019hardnet} based light-weight block for reducing the concatenation cost.
    Our enhancement network can be divided into two components: a Harmonic Dense Block (HBlock) for low memory computational cost and a Feature Identity Extraction Module (FIE) with feature fusion high-level perception task.
    
    \subsubsection{Harmonic Dense Block}
        To learn the recovery information, inspired by \cite{chao2019hardnet}, we model HBlock with depth $L$ layers.
        While the standard DenseNet passes the gradient from propagated all the layers, it leads to terrible large memory usage and heavy computational cost outweighing the gain.
        To solve these problems, the output of HBlock with depth-$L$ is acquired through concatenation with $L^{th}$ layer and all the previous odd-ordered layers.
        We also make the output of all even layers from 2 to $L$-2 to be removed once the HBlock is finished.
        Lastly, to adjust the dimension, we set the 32 channels in the last layer of each block.
        Each layer $L$ has an output channel width $k$, and the number of its channels is calculated by $k \times 1.6^{n}$, where $n$ is the maximum value at which the layer $l$ is divided by the integer quotient by $2^m$.
        
        Additionally, we employ a bottleneck layer before every 4th convolution layer to further accelerate the parameter efficiency, and set its output channel to $\sqrt{\frac{c_{in}}{c_{out}}}$, where $c_{in}$ and $c_{out}$ are channels of input and output, respectively.
        To this end, we propose two versions of the network, each consisting of 71 layers (5 HBlocks) and 33 layers (3 HBlocks).
        The batch normalization is used after each convolution layer except the last layer.
        After that, ReLU is applied as an activation function.
        Finally, in order to achieve more high-quality recovery, a recursive enhancement structure is introduced with a total of 3 stages, which is gradually leading to perception-friendly quality at the final stage.
        The full description of our enhancement network is shown in Table 1.
        
    \subsubsection{Feature Identity Extraction Module}
        The feature identity extraction module (FIE) is designed for correlating information from image enhancement and high-level visual perception features.
        The FIE is based on 3-layer CNN, which assigns exactly 128-dimensional latent features after the output of flattening the last layer of FIE with random projection instead of dense, as shown in Fig. 3.
        This allows unrestricted comparisons through random projection when the final layer output dimensions of FIE are different.
        Therefore, the FIE connects them by representing the mutual influence between image enhancement and visual perception in a unified framework.

\subsection{Training Strategy} 
    To learn the proposed network, we further integrate both image enhancement network and high-level network via three-stage.
    Our training strategy is divided into three parts: 1) image enhancement network learning, 2) high-level vision loss calculation, and 3) feature identity learning.
    
    \subsubsection{Image recovery loss}
        Existing state-of-the-art methods adopt the pixel-wise loss based on MSE (Mean-Squared error) to train enhancement network. 
        However, the MSE optimization usually produces blurry visual information which results in perceptual unsatisfactory images with over-smooth content. 
        To prevent this, we estimate the successive approximation to the bad weather distribution with the guidance of the Charbonnier penalty function \cite{lai2017deep}, which is more robust to outliers.
        The recovery loss is expressed as:
            \begin{equation}
            	{L_{R}(I^{pred}, I^{GT}) = || \sqrt{(I^{pred} - I^{GT})^2} ||_{2}^{2} + \epsilon^2},
            \end{equation}
        where $\epsilon$ is penalty coefficient and empirically set to $5\times10^{-3}$.
        We take one step further to give rich connectivity between the enhancement network and high-level perception.
    
    \subsubsection{High-level task loss}
        
        We use high-level task loss $L_{HT}$ from a pre-trained high-level vision task network to provide the enhancement network with connectivity that promotes it to be perception-friendly.
        By default, perception networks for high-level tasks are pre-trained on benchmarks composed of clean images and are frozen while learning the proposed framework.
        In addition, even if our enhancement network is replaced with another model, it can be replaced without any additional tuning to the coefficients of objective function and retraining of the perception network.
        As far as we know, this is the first study to run a variety of high-level tasks while dealing with all bad weather, taking a step further in universality.
        To convey more strong perception-friendly property, we describe the feature identity loss in the next.

    \subsubsection{Feature identity loss}
        \cite{johnson2016perceptual} propose to utilize a Euclidean distance which calculates identity information on image pairs, that proves to generate high-quality samples than the standard per-pixel losses. 
        Their idea has been adopted mostly in image generation work, such as super-resolution, translation, and image recovery.
        Despite the fact, we observed that even when the recognition tasks other than image generation is involved, the identity information is still essential for stable optimization.
        To give the relevant information in the training process, we propose to use a feature identity loss that leads to the directly related to identity in hypersphere space, defined as:
            \begin{equation}
            	{L_{FI}(f^{en}_{last}, f^{ht}_{last}) = || \reallywidehat{FIE(f^{en}_{last})} - \reallywidehat{FIE(f^{ht}_{last}) ||_{2}^{2}},}
            \end{equation}
        where $FIE(f^{en}_{last})$ and $FIE(f^{ht}_{last})$ are the identity features extracted from $(FIE)$ for input image $I^{X}$ and recovered image $I^{pred}$, respectively. $\reallywidehat{FIE(\cdot)}$ is the identity representation mapped to the hypersphere.
    
    \subsubsection{Objective Function}
        Based on the above introduction, we incorporate the above-mentioned losses as an objective function.
        We optimize the total objective function based on the stage-wise manner and can be trained by the following function:
            \begin{equation}
            \begin{split}
            	\min_{\{\theta_{en}, \phi\}}\frac{1}{N}\sum_{i=1}^{N}(L_{R}(I^{pred}_{i},I^{GT}_{i}) &+ \alpha L_{HT}(Y^{pred}_{i}, Y^{GT}_{i}) \\+ \beta L_{FI}(f^{en}_{last}, f^{ht}_{last})),
            \end{split}
            \end{equation}
        where $\alpha$ and $\beta$ are trade-off coefficients of the $L_{HT}$ and $L_{FI}$ respectively, $\theta_{en}$ and $\phi$ are learnable parameters from scratch with $N$ samples. 

\begin{table*}[]
\centering
\caption{\textbf{Quantitative bad weather enhancement evaluations with average PSNR/SSIM on synthetic images.} The best results in all methods are marked in bold. Second best are underlined.}
\label{tab:my-table}
\begin{tabular}{l|lllll}
\hline
Dataset & RESCAN \cite{li2018recurrent} & DeRaindrop \cite{qian2018attentive} & PReNet \cite{ren2019progressive} & Ours(33-layer) & Ours(71-layer) \\ \hline
KITTI & 22.95/0.7166 & 19.58/0.6845 & 25.19/0.7878 & \underline{25.56}/\underline{0.7932} & \textbf{27.06}/\textbf{0.8227} \\
TuSimple & 20.28/0.6581 & 20.97/0.7240 & 27.21/\underline{0.8266} & \underline{27.79}/0.8059 & \textbf{28.37}/\textbf{0.8348} \\ \hline
\end{tabular}
\end{table*}

\begin{table*}[]
\centering
\caption{\textbf{Quantitative high-level perception evaluations on two datasets (TuSimple, KITTI) and one real-world dataset RID.} Best results in each category are in bold. Second best are underlined.}
\vspace{-0.3cm}
\label{tab:my-table}
\begin{tabular}{ll|llllll}
\hline
 & metric & Bad Weather & RESCAN \cite{li2018recurrent} & DeRaindrop \cite{qian2018attentive} & PReNet \cite{ren2019progressive} & Ours(33-layer) & Ours(71-layer) \\ \hline
\multicolumn{1}{l|}{\multirow{3}{*}{Lane Detection}} & Acc $\uparrow$ & 95.86 & 95.16 & 95.38 & 95.54 & \underline{96.19} & \textbf{96.51} \\
\multicolumn{1}{l|}{} & FP $\downarrow$ & \textbf{2.85} & 5.35 & 4.69 & 5.15 & \underline{3.09} & 3.66 \\
\multicolumn{1}{l|}{} & FN $\downarrow$ & 3.70 & 5.59 & 5.04 & 4.79 & \underline{3.34} & \textbf{3.03} \\ \hline
\multicolumn{1}{l|}{\multirow{2}{*}{Depth Estimation}} & RMSE $\downarrow$ & 7.360 & 7.549 & 10.246 & 6.245 & \underline{5.501} & \textbf{5.351} \\
\multicolumn{1}{l|}{} & RMSE log $\downarrow$ & 0.343 & 0.322 & 0.487 & 0.262 & \underline{0.219} & \textbf{0.218} \\ \hline
\multicolumn{1}{l|}{Object Detection} & mAP $\uparrow$ & 23.92 & 21.87 & 22.51 & 23.80 & \underline{24.84} & \textbf{29.55} \\ \hline
\end{tabular}
\vspace{-0.2cm}
\end{table*}

    \begin{figure*}[t]
    \begin{center}
       \includegraphics[width=0.8\linewidth]{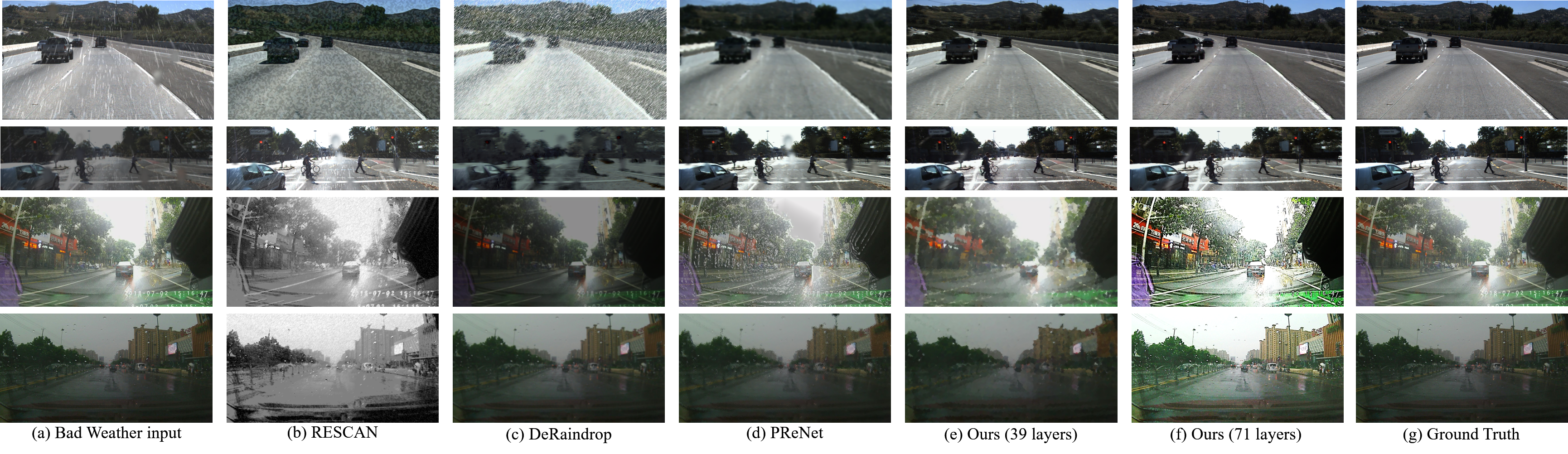}
    \end{center}
    \vspace{-0.6cm}
       \caption{\textbf{Visual comparison of different enhancement results.} For the four bad weather images in the first column (a), columns (b-d) show that the enhancement results by state-of-the-art methods, respectively. The proposed method contributes to getting better restoration results in column (e-f). Ground truth is shown in the last column (g). Best viewed on the computer, in color, and zoomed in.}
    \vspace{-0.2cm}
    \end{figure*}
    
    \begin{figure*}[t]
    \begin{center}
       \includegraphics[width=0.8\linewidth]{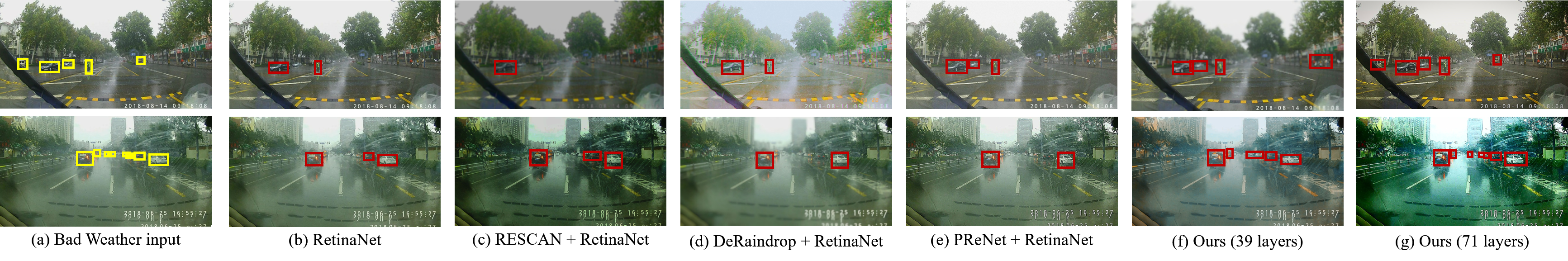}
    \end{center}
    \vspace{-0.6cm}
       \caption{\textbf{Visualization of object detection results on RID \cite{li2019single}.} The yellow bounding box on the bad weather input (a) is ground truth. Other bounding boxes in (b-g) are predicted results. Best viewed on the computer, in color, and zoomed in.}
    \vspace{-0.4cm}
    \end{figure*}
    
    \begin{figure*}[t]
    \vspace{1.3cm}
    \begin{center}
       \includegraphics[width=0.8\linewidth]{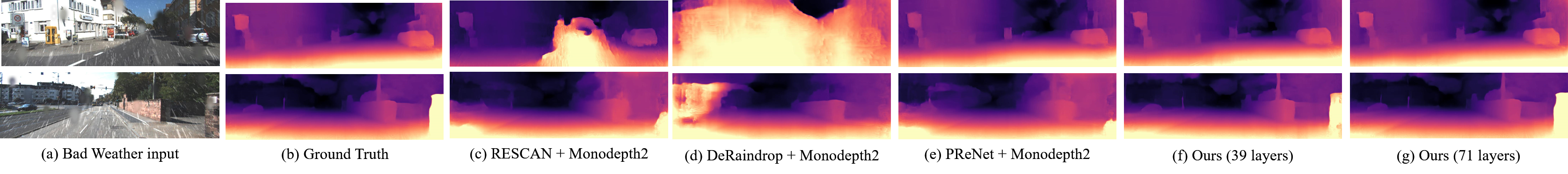}
    \end{center}
    \vspace{-0.3cm}
       \caption{\textbf{Visualization of monocular depth estimation results on KITTI \cite{geiger2012we}.} The results in (b) is the ground truth extracted from the clean image. Other results in (c-f) are predicted results from bad weather input (a). Best viewed on the computer, in color, and zoomed in.}
    \vspace{-0.3cm}
    \end{figure*}

    \begin{figure*}[t]
    \begin{center}
       \includegraphics[width=0.8\linewidth]{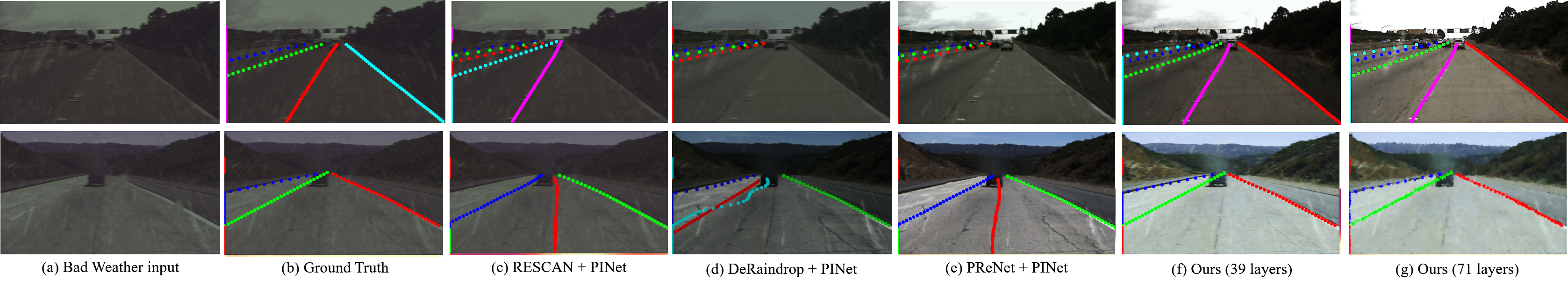}
    \end{center}
    \vspace{-0.5cm}
       \caption{\textbf{Visualization of lane detection results on TuSimple \cite{tuSimple}}. The results in (b) is the ground truth extracted from the clean image.Other results in (c-f) are predicted results from bad weather input (a). Best viewed on the computer, in color, and zoomed in.}
    \vspace{-0.5cm}
    \end{figure*}

\section{Experiments and Evaluation}
\vspace{-0.2cm}
\subsection{Datasets}
    In computer vision, few image datasets contain comprehensive bad weather conditions specific to driving situations, and likewise none of the datasets mentioned above are available.
    In order to create bad weather effects, we adopt rain streak, raindrop, and haze simultaneously on each image in the dataset (2 rain streaks $\times$ 1 raindrop $\times$ 2 haze effects).
    For the realistic rain streak effect, we are motivated from \cite{halder2019physics} and thus create two versions of streak intensities (heavy and light rain) with randomly distributed orientation.
    For the raindrop effect, we adopt a simulation method from \cite{ROLE} to apply the water drop on the lens to all images.
    For the haze effect, we employ widely used atmospheric scattering model introduced in \cite{mccartney1976optics,narasimhan2000chromatic} and generate two different hazy images under uniformly randomly chosen atmospheric lights and scattering coefficient as parameters.
    As a result, we obtain a paired dataset of clean target-bad weather images, where the split indexing of training and testing samples all follow the standard of the existing dataset. 
    
    For TuSimple dataset \cite{tuSimple} as lane detection benchmark, we take 3,626 images for training and 2,782 images for testing.
    For monocular depth estimation, we also take 39,810 images for training and 4,424 images for testing from KITTI benchmark \cite{geiger2012we}.
    Lastly, object detection evaluation of our model is performed on the RID dataset \cite{li2019single} which contains real bad weather such as rain streaks and densely disrupted haze, it only provides 2,495 samples for testing without synthetic effects.
    
    \vspace{-0.2cm}
\subsection{Training Configurations}
    The used datasets are resized to 1024 $\times$ 512 including both training and testing.
    All the networks are trained from scratch using Adam optimizer for 100 epochs with a total batch size of 8.
    The learning rate is first initialized to 0.0001 and divided by 5 following milestones at the 30$^{th}$, 50$^{th}$, and 80$^{th}$ epochs.
    The output channel width $k$ is [14,16,20,20,40] for the 71-layer and [14,16,40] for the 33-layer.
    Additionally, by empirical finding, the coefficient $\alpha$ for high-level tasks is set to 0.01, 0.05, and 0.002 for lane detection, depth estimation, and object detection, respectively.
    All the experiments are performed by using one NVIDIA TITAN X GPU and one Intel Core i7-6700K CPU based on the PyTorch framework.
     
    \vspace{-0.2cm}
\subsection{Experimental Configurations}
\vspace{-0.1cm}
    For evaluation of our model, we test the effectiveness of our method on three representative high-level tasks: lane detection, monocular depth estimation, and object detection.
    For the perception baselines, we employ state-of-the-art baselines: PINet \cite{ko2020key} for lane detection, Monodepth2 for monocular depth estimation \cite{godard2019digging}, and RetinaNet \cite{lin2017focal} for object detection.
    We also utilize their pre-trained weights from the publicly available codes.
    
    To quantitatively verify the effectiveness of the proposed method, we employ two types of metrics that measure task performance and image quality.
    For the image quality, we adopt PSNR and SSIM \cite{wang2004image}, which are standard \cite{cai2016dehazenet,tian2020deep} in image recovery.
    However, they may become loosely related when it comes to other high-level task purposes \cite{li2019single,pei2018does,sun2019convolutional}.
    Therefore, for the task-specific evaluation, we use standard metrics like the following: accuracy for lane detection, RMSE for monocular estimation, and mAP for object detection.
    Note that the proposed setting is evaluated without requiring manual data annotation in a comprehensive and fair setting.

\begin{table}[]
\vspace{-0.3cm}
\centering
\caption{\textbf{Computational cost analysis.}}
\vspace{-0.3cm}
\label{tab:my-table}
\begin{tabular}{l|lll}
\hline
models & image size & FLOPs (G) & params (M) \\ \hline
RESCAN & 1024x512 & 258.57 & 0.15 \\
DeRaindrop & 1024x512 & 716.39 & 6.24 \\
PReNet & 1024x512 & 531.51 & 0.16 \\
Ours (33 layers) & 1024x512 & \textbf{57.02} & \textbf{0.11} \\
Ours (71 layers) & 1024x512 & \underline{146.16} & \underline{0.28} \\ \hline
\end{tabular}
\vspace{-0.6cm}
\end{table}

\subsection{Image Enhancement Evaluations}
    Table 2 shows that our method achieves significant gains in terms of both PSNR and SSIM.
    As shown in Fig. 4, qualitative results reveal great effectiveness, while the result by DeRaindrop still contains visible bad weather elements.
    Consequently, the visual quality enhancement by our methods is significant, while the results by existing methods still contain visible bad weather effects.
    Moreover, our light-weight model (33-layer) shows PSNR on par with the other three methods.
    To our best knowledge, our 71-layer model is the only bad weather enhancement method so far that can simultaneously achieve clean visual quality and hardware-friendly property.
    Such a large gain demonstrates our method generates promising image enhancement results, which are visually more clear. 
    
\vspace{-0.3cm}
\subsection{High-Level Task Evaluations}
    To evaluate the effectiveness of our enhancement framework on high-level tasks in bad weather, we further show that the method yields meaningful results.
    From Table 3, our method outperforms the high-level perception performances in comparison for different methods using RESCAN, DeRaindrop, and PReNet as a pre-processing.
    As reported in \cite{li2019single}, most models do not improve over the bad weather input in terms of high-level perception metrics.
    Instead, our method is observed to improve performance in high-level tasks without deteriorating.
    This proves that our method represents mutual influence between visual perception and enhancement, thus providing significant help for bad weather capabilities.
    Fig. 5, 6 and 7 also show that our proposed method outperforms the existing methods, confirming the perception-friendly ability of the model to bad weather.
    
    Table 4 reports the computational cost of our enhancement network and some state-of-the-art methods.
    From the results, we can find that our method has less computational overload due to the harmonic dense block.
    Taking their hardware-friendly ability into account, it is appealing to still maintain perception performance when facing bad weather images.
    
\begin{table}[]
\centering
\caption{\textbf{Ablation Study.} HLT refers to the baselines corresponding to the task. T1 to 3 refer to lane detection, monocular depth estimation, and object detection, respectively.}
\vspace{-0.3cm}
\label{tab:my-table}
\begin{tabular}{lll|llll}
\hline
 & metric & model & (a) HLT & \begin{tabular}[c]{@{}l@{}}(b) +EN\\ (w/o\\ training)\end{tabular} & \begin{tabular}[c]{@{}l@{}}(c) +EN\\ (training)\end{tabular} & \begin{tabular}[c]{@{}l@{}}(d) +FIE\\ (Ours)\end{tabular} \\ \hline
\multicolumn{1}{l|}{\multirow{2}{*}{T1}} & \multirow{2}{*}{Acc} & 71L & 95.86 & 94.23 & 96.37 & \textbf{96.51} \\
\multicolumn{1}{l|}{} &  & 33L & 95.86 & 95.43 & 95.95 & \textbf{96.19} \\\hline
\multicolumn{1}{l|}{\multirow{2}{*}{T2}} & \multirow{2}{*}{RMSE} & 71L & 7.360 & 5.595 & 5.428 & \textbf{5.351} \\
\multicolumn{1}{l|}{} &  & 33L & 7.360 & 5.616 & 5.556 & \textbf{5.501} \\\hline
\multicolumn{1}{l|}{\multirow{2}{*}{T3}} & \multirow{2}{*}{mAP} & 71L & 23.92 & 22.31 & 26.84 & \textbf{29.55} \\
\multicolumn{1}{l|}{} &  & 33L & 23.92 & 22.84 & 23.75 & \textbf{24.84} \\ \hline
\end{tabular}
\vspace{-0.5cm}
\end{table}

\subsection{Ablation Studies}
     To study the contribution of each network in our proposed framework, we alternatively remove it and identify the impact on the high-level perception performance.
     As can be seen in Table 5, our method with joint training (c-d) performs better than simple connection (b).
     Joint training (c) has a slightly lower baseline in Task1 and Task3, which is not surprising since it was not trained with the feature identity network.
     After applying the FIE (d), our method is the best-ranked approach that significantly outperforms the other three options (a-c), and we are encouraged to observe that the FIE brings the interconnection between the two networks.
     Finally, it is observed that all the networks in our proposed framework lead to important contribution in the final performance.


\vspace{-0.2cm}
\section{CONCLUSIONS}
    In this paper, we have proposed a novel task-driven image enhancement framework connected to visual perception for autonomous driving under the presence of bad weather.
    In particular, we have revealed that the existing methods are not practical for real-world autonomous driving in resource-constrained devices, and have aimed to improve them from two perspectives.
    First, our method is \textit{perception-friendly} since it is optimized not only for the human-centric visibility but also for the high-level task models simultaneously.
    In addition, we developed a low-memory network architecture, focusing on a \textit{hardware-friendly} ADAS system on the embedded system suitable for autonomous cars.
    Compared to previous methods, our method has verified improved performance in terms of both perception and hardware for autonomous driving despite bad weather.
    Future work will focus on modeling bad weather characteristics explicitly to remove artifacts and preserve details more effectively.

\addtolength{\textheight}{-12cm}   







\addtolength{\textheight}{10cm}
\bibliography{egbib}
\bibliographystyle{ieeetran}

\end{document}